\documentclass[10pt,twocolumn,letterpaper]{article}

\usepackage{iccv}
\usepackage{times}
\usepackage{epsfig}
\usepackage{graphicx}
\usepackage{amsmath}
\usepackage{amssymb}

\usepackage{multirow}

\usepackage[breaklinks=true,bookmarks=false]{hyperref}

\iccvfinalcopy 

\def\iccvPaperID{****} 

\ificcvfinal\fi

\begin{document}

\title{Analysis of Deep Image Quality Models}

\author{Pablo Hernández-Cámara\\
Image Processing Lab\\
Universitat de València, Spain\\
{\tt\small pablo.hernandez-camara@uv.es}
\and
Jorge Vila-Tomas\\
{\tt\small jorge.vila-tomas@uv.es}
\and
Valero Laparra\\
{\tt\small valero.laparra@uv.es}
\and
Jesús Malo\\
{\tt\small jesus.malo@uv.es}
}

\maketitle
\ificcvfinal\thispagestyle{empty}\fi

\begin{abstract}

Subjective image quality measures based on deep neural networks are very related to models of visual neuroscience.
This connection benefits engineering but, more interestingly, the freedom to optimize deep networks in different ways, make them an excellent tool to explore the principles behind visual perception (both human and artificial). 
Recently, a myriad of networks have been successfully optimized for many interesting visual tasks. Although these nets were not specifically designed to predict image quality nor other psychophysics, they have shown surprising human-like behavior. The reasons for this remain unclear. 

In this work, we perform a thorough analysis of the perceptual properties of pre-trained nets (in particular their ability to predict image quality) by isolating different factors: 
the goal (the function), the data (the learning environment), 
the architecture, and the readout:  selected layer(s), fine-tuning of channel relevance, and use of statistical descriptors as opposed to plain readout of responses. 

Several conclusions can be drawn. 
All the studied models correlate better with human opinion than SSIM (a \emph{de-facto} standard). More importantly, some of the nets are in pair of the state-of-the-art with no extra refinement nor perceptual information.
Nets trained for supervised tasks such as classification correlate substantially better with humans than LPIPS (a recent net specifically tuned for image quality). Interestingly self-supervised tasks such as jigsaw also perform better than LPIPS.
Simpler architectures are better than very deep nets. 
In simpler nets, correlation with humans increases with depth as if deeper layers were closer to human judgement.
This is not true in very deep nets.
Consistently with reports on illusions and contrast sensitivity, small changes in the image environment does not make a big difference in performance. Finally, the explored statistical descriptors and concatenations had no major impact.

\end{abstract}

\section{Introduction}

Humans have evolved to adapt our visual system to the statistics of nature. Neural networks learn the statistics of the data they have been trained on \cite{iclr_2022}. However, how human perception and machine learning models trained with natural images are related is an open issue. There is a long tradition on relating human visual system with machine learning models and  there are many options to relate them. One option is trying to reproduce fundamental signatures of the visual system that have been directly measured, such us contrast sensitivity functions (CSF) \cite{csf_autoencoders, arash_gegenfurtner}. Other option that we explore here is the image quality assessment (IQA) problem, which tries to predict image distances in a similar way to what a human would do.

\begin{table*}
\begin{center}
\begin{tabular}{c|c|c|c|c}
Architecture & Goal & Training data & Read-out & Fine tuning \\
\hline\hline
AlexNet & \multirow{5}{*}{Supervised} & \multirow{5}{*}{ImageNet-1K} & \multirow{5}{*}{Euclidean, no concat.} & \multirow{5}{*}{No} \\
VGG-16 &  &  &  &  \\
ResNet-50 &  &  &  &  \\
DenseNet-121 &  &  &  &  \\
EfficientNet-B0 &  &  &  &  \\
\hline
\multirow{5}{*}{AlexNet} & Supervised & \multirow{5}{*}{ImageNet-1K} & \multirow{5}{*}{Euclidean, no concat.} & \multirow{5}{*}{No} \\
 & Self-Supervised RotNet &  &  &  \\
 & Self-Supervised Jigsaw &  &  &  \\
 & Self-Supervised Colorization &  &  &  \\
 & Self-Supervised DeepCluster &  &  &  \\
 \hline
\multirow{3}{*}{AlexNet} & \multirow{3}{*}{Supervised} & ImageNet-1K & \multirow{3}{*}{Euclidean, no concat.} & \multirow{3}{*}{No} \\
 &  & Places-365 &  &  \\
  &  & Cifar-10 &  &  \\
 \hline
\multirow{8}{*}{AlexNet} & \multirow{8}{*}{Supervised} & \multirow{8}{*}{ImageNet-1K} & Euclidean, no concat. & \multirow{8}{*}{No} \\
 &  &  & Euclidean, concat. &  \\
 &  &  & Means, no concat. &  \\
 &  &  & Means, concat. &  \\
 &  &  & Means-sigmas, no concat. &  \\
 &  &  & Means-sigmas, concat. &  \\
 &  &  & Gram, no concat. &  \\
 &  &  & Gram, concat. &  \\
 \hline
\multirow{3}{*}{AlexNet} & \multirow{3}{*}{Supervised} & \multirow{3}{*}{ImageNet-1K} & \multirow{3}{*}{Euclidean} & No \\
 &  &  &  & TID-2008 \\
 &  &  &  & train-KADID-10K
\end{tabular}
\end{center}
\caption{Resume of the factors analyzed. Each column show the different model factors that we have varied and each row shows the different options explored.}
\label{table_article}
\end{table*}

A classical tradition to predict human perception is to use biologically plausible models to define the quality metrics \cite{perceptnet}. However, this is not mandatory and it is possible to approach the problem from other perspectives. For many years, statistical based quality methods have been widely used and awarded \cite{ssim}. In the last decade, features extracted from deep learning models have been used to calculate the distances without taking care of the biological plausibility of the architecture (LPIPS \cite{lpips} and DISTS \cite{dists}). These models use convolutional feature extractors (pre-trained to perform classification) as a backbone to define image quality metrics in their inner domain and perform a fine-tune with few perceptual data to maximize the correlation with human perception. Results from these models correlate surprisingly well with human perception and they have become state-of-the-art in image quality assessment. However, the reason why these models designed for other tasks correlate so well with human perception remains unclear.

In recent years, some studies tried to analyze different aspects of this relation between machine learning models and human perception \cite{iclr_2022, csf_autoencoders, arash_gegenfurtner, papergoogle}. They found, not only that low level tasks have better properties than high level tasks for reproduce human-CSF, but also that models with higher accuracy on ImageNet classification perform worse when correlating with human perception. However, there are still many open issues to clarify these behaviors. 

In this work, we do an analysis of the perceptual properties of already trained models when facing image quality problems. We analyze how different factors such as architecture, objective, data, and ways of computing the distance (output deepness, output concatenation, channel relevance fine-tuning, and using statistics) affect the correlation between human perception and model distances. Table \ref{table_article} shows a summary of the different tested factors. To the best of our knowledge, this is the first work that analyzes how well the features extracted (layer by layer) correlate with human perception.

\section{Related work}

Since its appearance in 2004, SSIM became the standard model for predicting image distances as close to humans as possible \cite{ssim}. It is based on calculating statistical descriptors of image patches to compare them an obtain a distance measurement. However, SSIM recently was surpassed by models based on pre-trained artificial neural networks. For example, LPIPS \cite{lpips} concatenates features of different VGG layers. Once an image and its distorted version have passed through the network and the outputs have been read, each feature is weighted to maximize the correlation with human perception. Even fewer years ago, in 2020, DISTS \cite{dists} unified both SSIM and LPIPS. It uses different layers of VGG to extract features from the images, and computes statistical descriptors as in SSIM to compare them in order to get a distance measurement between an image and its distorted version.

These IQA models based on neural networks correlate surprisingly well with human perception, although the reason is not known. In the last years, some works have tried to explore and understand this relation between deep learning models and human perception. On the one hand, some studies studied fundamental signatures of human visual system to check if deep learning models also have the same signatures. They show that human-CSFs appear more in low/middle-level tasks than in high-level objectives \cite{arash_gegenfurtner, csf_autoencoders}. They also show that depending on the training objective, CSFs appear more in some layers than in others. This shows that, as expected, neither all training targets nor all layers of deep learning models have the same properties, and that some of them are more closely related to human perception than others.

On the other hand, other works explore the relation between deep learning models classification accuracy and human perception. They show that there is an inverse-V relation between ImageNet accuracy and correlation with human perception in IQA problems \cite{papergoogle}, where models with high and very low accuracies have lower correlation with humans. 

\section{Method}

Here we describe the databases and models we used in our experiments and the different ways we employed to calculate distances between images in the inner representation of a certain model.

\subsection{Databases and models}

We restrict ourselves to already trained models by third-party people in order to avoid dependence on architecture design or training procedures. Following this, we used different pre-trained models, trained on ImageNet \cite{deng2009imagenet}, Places-365 \cite{places365} and Cifar-10 \cite{cifar10} databases, in supervised or self-supervised ways. More particularly, for the supervised models, we used AlexNet \cite{alexnet}, VGG-16 \cite{vgg16}, DenseNet-121 \cite{densenet}, ResNet-50 \cite{resnet} and EfficientNet-B0 \cite{efficientnet}. We downloaded AlexNet and EfficientNet-B0 from TorchVision \cite{torchvision} and VGG-16, DenseNet-121 and ResNet-50 from Keras \cite{chollet2015keras}. We downloaded all of them with ImageNet pre-trained weights. We also downloaded the supervised AlexNet model pre-trained in Places-365 from MIT CSAIL Computer Vision \cite{alexnet_places} and pre-trained in Cifar-10 from \cite{alexnet_cifar}. For the self-supervised models, we select one architecture (AlexNet) and use different self-supervised training goals with ImageNet images. We tested RotNet \cite{rotnet}, Jigsaw \cite{jigsaw}, Colorization \cite{colorization} and DeepCluster \cite{deepcluster} tasks. We downloaded all the self-supervised models from Facebook research VISSL \cite{goyal2021vissl}. Table \ref{table_models} shows a resume of the supervised and self-supervised models trained with ImageNet-1K data and their ImageNet-1k Top 1 accuracy. 

\begin{table}
\begin{center}
\begin{tabular}{c|c|c}
Architecture & Training process & ImageNet Top 1\\
\hline\hline
AlexNet & Supervised & 56.5\% \\
VGG-16 & Supervised & 71.3\% \\
DenseNet-121 & Supervised & 75.0\% \\
ResNet-50 & Supervised & 74.9\% \\
EfficientNet-B0 & Supervised & 77.7\% \\
\hline
AlexNet & Self RotNet & 39.5\% \\ 
AlexNet & Self Jigsaw & 34.8\% \\ 
AlexNet & Self Colorization & 30.4\% \\ 
AlexNet & Self DeepCluster & 37.9\% \\ 
\end{tabular}
\end{center}
\caption{Summary of the tested models, their training process and their ImageNet Top 1 accuracy.}
\label{table_models}
\end{table}

To test the models described previously we used two image quality databases: TID-2013 \cite{PONOMARENKO_tid} and KADID-10K \cite{kadid10k}. Both consists on pairs of images and distorted versions of the same image with a mean opinion score (MOS) for each image - distorted image pair, which represents the distance between them as estimated by humans. More particularly, we used the whole TID-2013 (3000 image pairs) and $30\%$ of KADID-10K, which we call val-KADID-10K (3038 image pairs), for testing the different models.

We also tested weighting the features extracted by the models (what we call fine-tuning), similarly to LPIPS. We used TID-2008 (1700 image pairs) and the remaining $70\%$ of KADID-10K (train-KADID-10K) (7087 image pairs) to weight each output feature in order to maximize the correlation with the MOS in these training databases.

\subsection{Distance measurement and correlation}

There are several ways to measure the distance between two images. For, example, let $x_{0} \in \mathbb{R}^{(H,W,C)}$ and $y_{0} \in \mathbb{R}^{(H,W,C)}$ denote an image and its distorted version, and $\hat{x}^{l} \in \mathbb{R}^{(H^l,W^l,C^l)}$ and $\hat{y}^{l} \in \mathbb{R}^{(H^l,W^l,C^l)}$ their feature maps at the $l^{th}$ layer of a network. Then, their euclidean distance is just:

\begin{equation}
d^{l}(x_{0}, y_{0}) = \sqrt{\sum_{H^{l}, W^{l}, C^{l}}(\hat{x}^{l} - \hat{y}^{l})^2}
\end{equation}

However, it is also possible to statistically summarize the layer output before computing the distance. Here, we used three different ways of summarize the layer output. First, we can use the mean of each feature:

\begin{equation}
d^{l}_{\mu}(x_{0}, y_{0}) = \sqrt{\sum_{C^{l}}(\hat{\mu}^{l}_{x} - \hat{\mu}^{l}_{y})^2}
\end{equation}

where $\hat{\mu}^{l}_{x} \in \mathbb{R}^{C^l}$ and $\hat{\mu}^{l}_{y} \in \mathbb{R}^{C^l}$ are the spatial average of the outputs of the $l^{th}$ layer for the image $x_{0}$ and its distorted version $y_{0}$. In this case, we compute the spatial averages to calculate the distance with them. Second, we can use not only the spatial averages but also the spatial standard deviations. In this case, we concatenate the spatial averages and standard deviation before computing the distance:

\begin{equation}
d^{l}_{\mu,\sigma}(x_{0}, y_{0}) = \sqrt{\sum_{C^{l}}(concat(\hat{\mu}^{l}_{x},\hat{\sigma}^{l}_{x}) - concat(\hat{\mu}^{l}_{y},\hat{\sigma}^{l}_{y}))^2}
\end{equation}

where $\hat{\mu}^{l}_{x} \in \mathbb{R}^{C^l}$, $\hat{\mu}^{l}_{y} \in \mathbb{R}^{C^l}$ and $\hat{\sigma}^{l}_{x} \in \mathbb{R}^{C^l}$, $\hat{\sigma}^{l}_{y} \in \mathbb{R}^{C^l}$ are the spatial average and standard deviation of the outputs of the $l^{th}$ layer for the image $x_{0}$ and its distorted version $y_{0}$ respectively. Finally, we can summarize the outputs through their Gram matrix:

\begin{equation}
d^{l}_{G}(x_{0}, y_{0}) = \sqrt{\sum_{C^{l}}(\hat{G}^{l}_{x} - \hat{G}^{l}_{y})^2}
\end{equation}

where $\hat{G}^{l}_{x} \in \mathbb{R}^{(C^l, C^l)}$ and $\hat{G}^{l}_{y} \in \mathbb{R}^{(C^l, C^l)}$ are the Gram matrices of the outputs at the $l^{th}$ layer for the image $x_{0}$ and its distorted version $y_{0}$.

Inspired by some of the SOTA image quality models \cite{lpips}, one can also concatenate the outputs of different layers in order to introduce more information to calculate the distance. Not only that, it is also possible to weight (fine-tune) the output features of a model (concatenating or not) so that the correlation with a specific database is maximized.


Summing up, to obtain the results we pass each image - distorted image pair through the different models and record the outputs at different layers. Then, we calculate the distance between them at the different layers using one of the distance definitions from above ($d^{l}$-$d^{l}_{\mu}$-$d^{l}_{G}$; concatenating or not the outputs of different layers; weighting or not the outputs). Finally, we calculate the Spearman correlation between the distance of the model at different layers with the experimental MOS.

\section{Experiments and results}

\subsection{Architecture}

In our first experiment, we tested how different architectures (AlexNet, VGG-16, ResNet-50, DenseNet-121 and EfficientNet-B0) correlate with human perception. Here we fixed the goal of the models (ImageNet supervised classification task), the training data (ImageNet-1K), the way we perform the read-out to calculate the distance (euclidean: $d^{l}(x_{0}, y_{0})$) and we did not weight the output features, so we just calculate the euclidean distance at the inner domain of each layer. 

\begin{figure*}
\begin{center}
\includegraphics[width=0.95\linewidth]{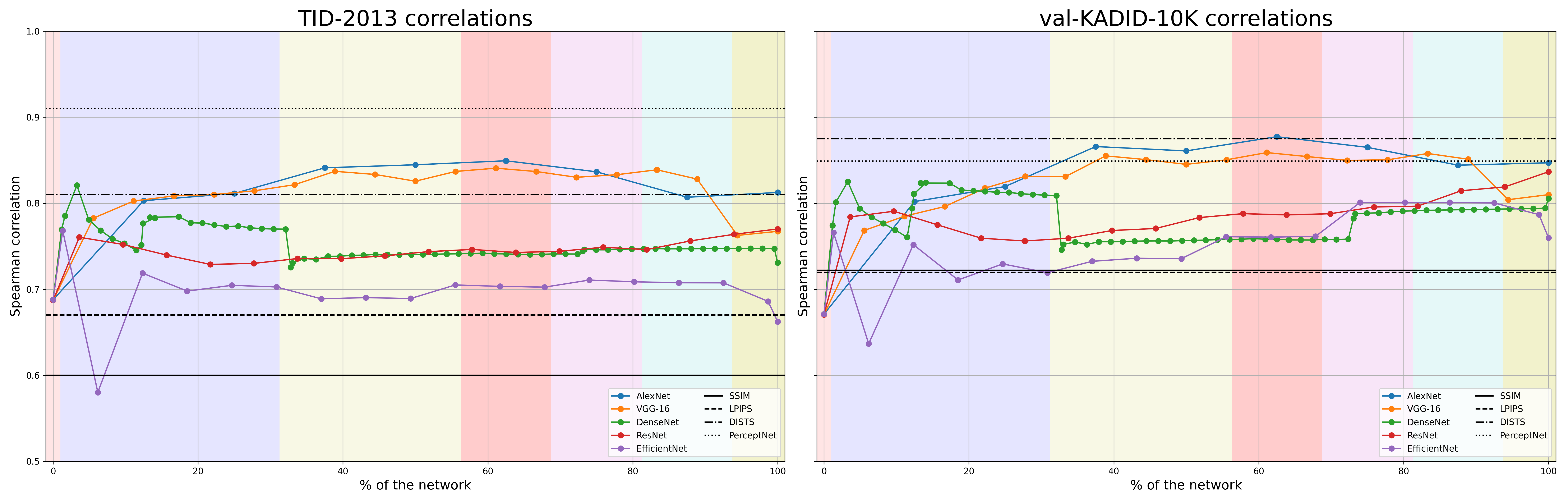}
\end{center}
   \caption{TID-2013 (left) and val-KADID-10K (right) Spearman correlation at different model depths (different layers) for different model architectures. Note that each model has different number of layer so that to plot all together, the x-axis represents the percentage of the network. Colors represent the different AlexNet blocks, which achieve the higher correlation at the second half of the network. Some published IQA models results are shown in black (solid and dashed lines).}
\label{fig:architecture}
\end{figure*}

Figure \ref{fig:architecture} shows how different architectures correlate with human perception (MOS) at different depths (different layers) for TID-2013 and val-KADID-10K.

There are several results to notice from this figure. First, all the models perform better than classical statistical image quality models (SSIM \cite{ssim}: black solid line). Also, simpler models (AlexNet and VGG-16) perform better than modern image quality algorithms based on neural networks (DISTS \cite{dists}, LPIPS \cite{lpips}: black dashed lines). Only some specific biologically inspired image quality algorithms \cite{perceptnet} perform better than the majority of the models in TID-2013. Second, simpler models achieve higher correlations. Specifically, AlexNet and VGG-16, which do NOT have skip connections, get more correlation with human perception than modern models with higher ImageNet accuracies. The fact that skip connections hurt the models is in agreement with the results of style transfer, where it was found that using ResNet with skip connections performs worse than networks without skip connections such as VGG \cite{bad_skip_connections}. Third, in the simplest models (AlexNet and VGG-16), there is a relationship between the depth of the layer used to measure image distances and the correlation with perception, with higher correlations obtained in deeper layers. However, both models show a decrease in correlation for the last two layers. More complicated models (ResNet-50, DenseNet and EfficientNet) have much more complex correlation diagrams with depth, not showing a clear relation between deepness and correlation.

This suggests that image quality algorithms that used deep learning models should use classical networks without skip connections even if they do not achieve high accuracy in ImageNet classification.

\subsection{Goal function}

In the last years, some studies showed that human-CSF emerge more in low level tasks models such as denoising autoencoders \cite{csf_autoencoders, arash_gegenfurtner}. However, what about self-supervised tasks? Here we select the network architecture that obtained the higher correlation in the supervised scenario, AlexNet, and check if some self-supervised goals achieve higher correlation with human perception. To do this, we fixed the architecture (AlexNet), the training data (ImageNet-1K), the way we perform the read-out to calculate the distance (euclidean, $d^{l}(x_{0}, y_{0})$) and we did not weight the features, so that we just calculate the euclidean distance in the inner domain of each layer for AlexNet trained for different goals. 

\begin{figure*}
\begin{center}
\includegraphics[width=0.95\linewidth]{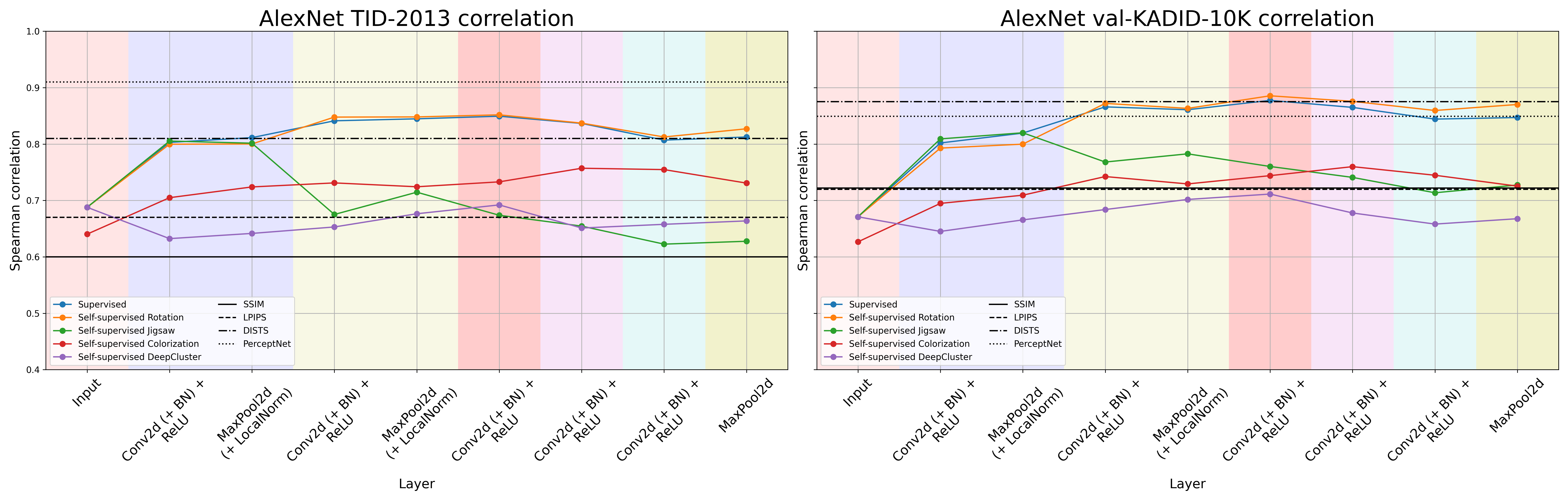}
\end{center}
   \caption{TID-2013 (left) and val-KADID-10K (right) Spearman correlation at different model depths (different layers) for different goals. Colors represent the different AlexNet blocks. Some IQA models results are shown in black solid and dashed lines.}
\label{fig:goal}
\end{figure*}

Figure \ref{fig:goal} shows how AlexNet trained with different objectives correlates with human perception (MOS) at different layers for TID-2013 and val-KADID-10K.

While all training objectives have good perceptual properties, there are some that have better properties than others. In fact, all the models achieve higher correlation than SSIM. We obtain that the supervised model and RotNet model obtain the higher correlations. However, it is important to highlight that the RotNet model was pre-trained on the supervised ImageNet task. Between the other self-supervised tasks, Jigsaw obtains correlations at the level of the supervised model only in its first layers but, as depth increases, the correlation goes down. Colorization goal shows a small linear increase in correlation with layer depth but it always has lower correlation than the supervised one. Finally, DeepCluster model remains always almost at the same correlation level than in the RGB (input) domain, which is just the correlation with the RMSE.

This suggests that image quality algorithms based on deep learning models should not use models trained for self-supervised tasks, but rather they should use supervised or low-level tasks.

\subsection{Training data and learning environment}

\begin{figure*}
\begin{center}
\includegraphics[width=0.95\linewidth]{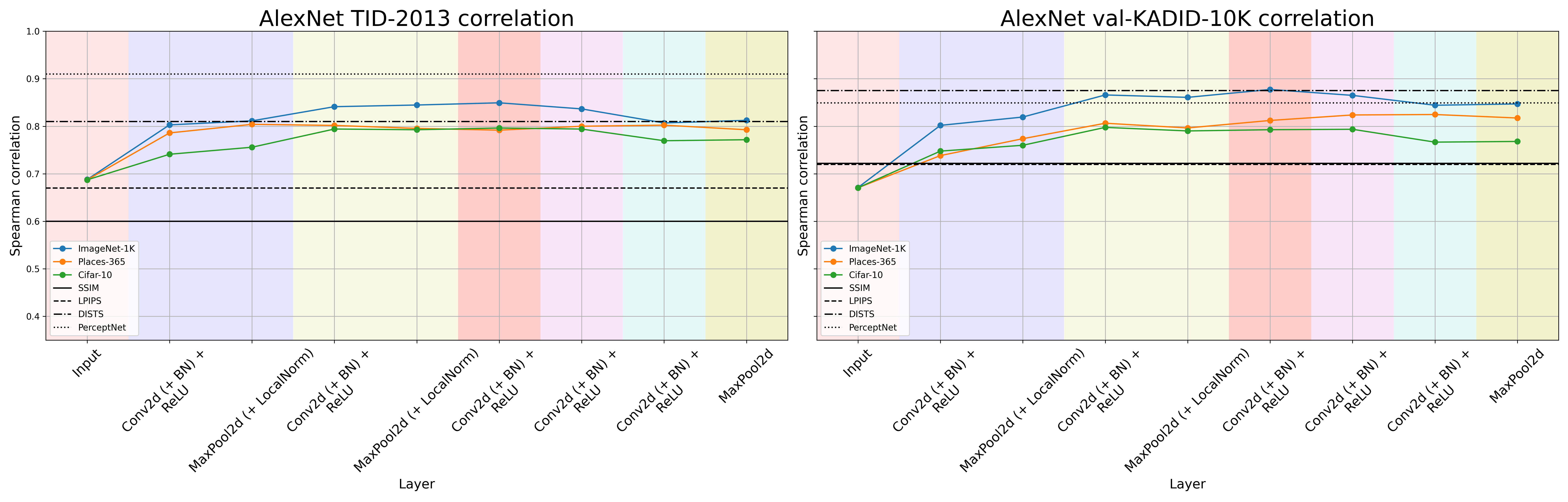}
\end{center}
   \caption{TID-2013 (left) and val-KADID-10K (right) Spearman correlation at different model depths (different layers) for different training data. Colors represent the different AlexNet blocks. Some IQA model results are shown in black solid and dashed lines.}
\label{fig:data}
\end{figure*}

We check how human perception correlates with the distances from networks trained with different data. We fixed the architecture that obtained a better correlation (AlexNet) and the training objective, supervised. We also keep fixed the way we perform the read-out to calculate the distance (euclidean, $d^{l}(x_{0}, y_{0})$) and we did not weight the features. The only difference between them is data used to train the models, and we analyze how the correlation depends on training on ImageNet-1k, Places-365 and Cifar-10. Figure \ref{fig:data} shows how different different training data correlate with human perception (MOS) at different layers for TID-2013 and val-KADID-10K.

There are not big differences between the data used in the training process. However, the best result is obtained with ImageNet-1K, which are around 1 million of natural images. Places-365 (ten million place locations images) and Cifar-10 (50K small natural images) achieve less correlation. It implies that not using natural images or using smaller/not enough images doesn't hurt the model too much.

It suggests that image quality algorithms using deep learning models should use as natural and big images as possible, while the data quantity isn't as important.

\subsection{Readout strategies and statistical descriptors}

\begin{figure*}
\begin{center}
\includegraphics[width=0.95\linewidth]{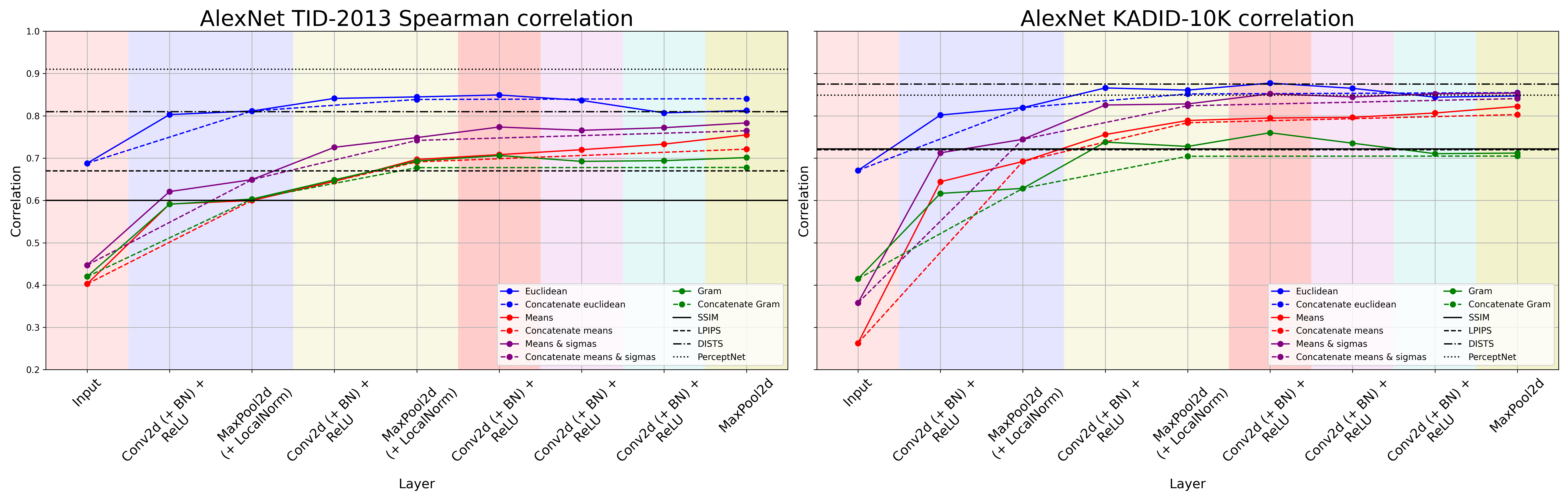}
\end{center}
   \caption{TID-2013 (left) and val-KADID-10K (right) Spearman correlation at different model depths (different layers) for different options to calculate the distance: just the euclidean distance with the full output, using statistical descriptors and/or concatenating the outputs of different layers. Colors represent the different AlexNet blocks. Some IQA model results are shown in black solid and dashed lines.}
\label{fig:read_out}
\end{figure*}

Here we tested how the way we perform the read-out affects the correlation. To test it, we use the supervised AlexNet model trained in a supervised way with ImageNet-1K and we calculate the correlation with human perception using the different ways to calculate the distances in the model inner domain. We tested not only the different distance definitions (summarizing the layer outputs with statistics or not) but also we check what happens when we concatenate the outputs of the three max pooling layers. Figure \ref{fig:read_out} shows how different distance measurements correlate with human perception (MOS) at different layers for TID-2013 and val-KADID-10K.

First, figure \ref{fig:read_out} shows that the best correlation is obtained when the whole output is used to calculate the distances, without the use of any statistical descriptor. When statistical descriptors are used, using the spatial Gram Matrix performs worst. Using the spatial mean together with the spatial standard deviation lead to higher correlation than using only the spatial means, because it implies using more information to calculate the distances. With regard to concatenating different layer outputs (such as in LPIPS \cite{lpips} and DISTS \cite{dists}), it does not have a big effect.  

This suggests that image quality algorithms that use deep learning models should utilize the full output of the layers without using any statistical descriptor. Also, using a concatenation of the outputs of different layers seems to have no benefits while increasing the computational complexity.

\subsection{Fine-tuning strategies.}

\begin{figure*}
\begin{center}
\includegraphics[width=0.95\linewidth]{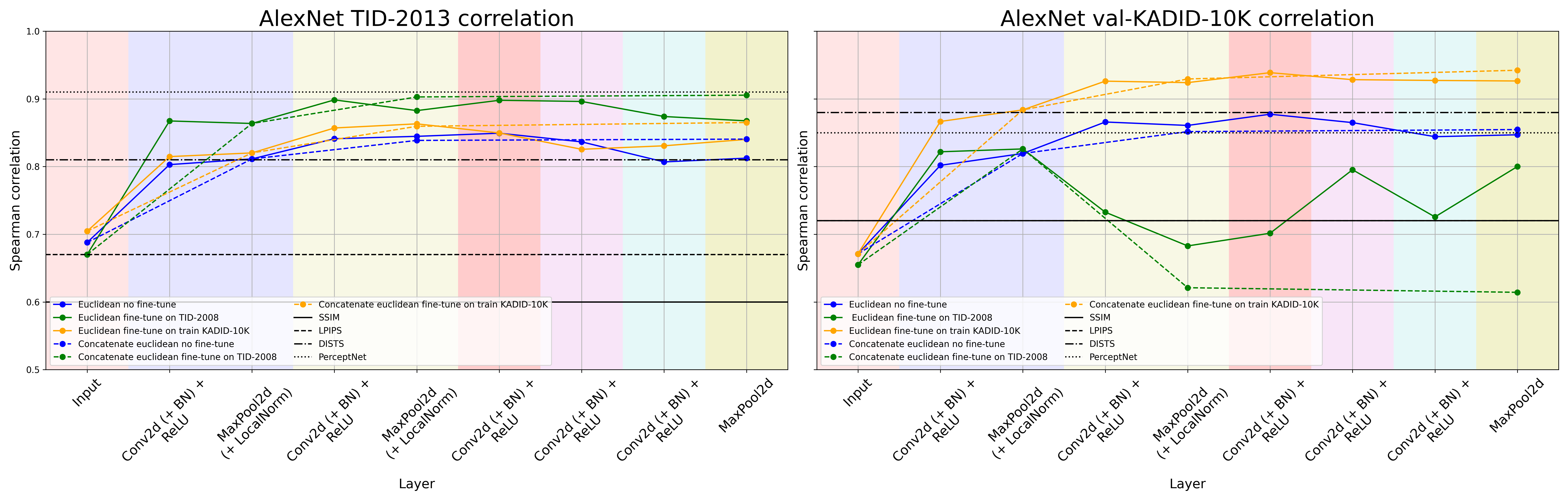}
\end{center}
   \caption{TID-2013 (left) and val-KADID-10K (right) Spearman correlation at different model depths (different layers) for different fine-tune, i.e. different feature weights. Colors represent the different AlexNet blocks. Some IQA model results are shown in black solid and dashed lines.}
\label{fig:fine_tune}
\end{figure*}

Finally, we perform an analysis of what happen when we weight each feature to maximize the correlation in some database. In this scenario we take the model that gives the higher correlation (AlexNet) trained in a supervised way on ImageNet-1K. We used only the euclidean read-out without any statistical descriptor measurement with and without concatenating the layer outputs. Figure \ref{fig:fine_tune} shows how different ways of weight features correlate with human perception. More specifically, we fine-tune the outputs with TID-2008 and train-KADID-10K and we tested at different layers with TID-2013 and val-KADID-10K.

The first conclusion is that fine-tuning with TID-2008 when evaluating in TID-2013 results in an increase in the correlation, which makes sense because the images are similar. It also occurs when fine-tuning with train-KADID-10K and evaluating with val-KADID-10K. More interesting results happened when we performed cross fine-tune: fine-tuning with train-KADID-10K gives no substantially changes when evaluating with TID-2013. However, fine-tuning with TID-2008 gives much worse results when evaluating in val-KADID-10K.

\subsection{Relation between ImageNet-1K classification accuracy and perceptual correlation}

\begin{figure*}
\begin{center}
\includegraphics[width=0.95\linewidth]{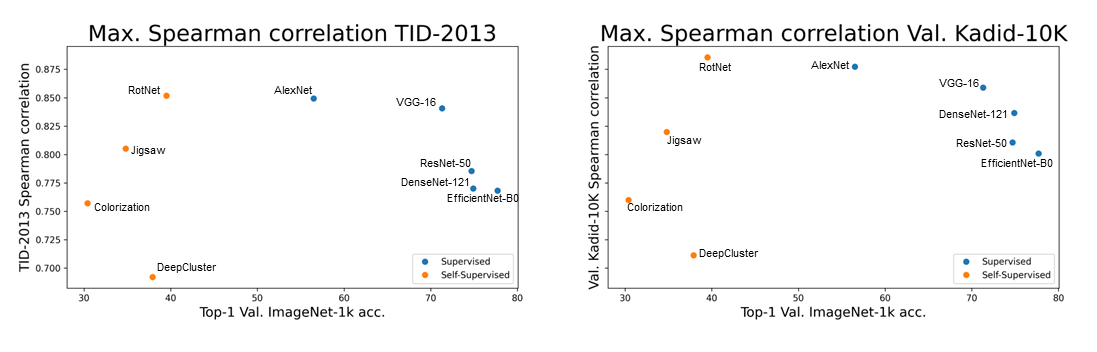}
\end{center}
   \caption{TID-2013 (left) and val-KADID-10K (right) maximum Spearman correlation relation with ImageNet-1K classification accuracy for the different models.}
\label{fig:v_invertida}
\end{figure*}

Finally, here we tried to compare the ImageNet-1K classification correlation with the maximum correlation obtained both in TID-2013 and val-KADID-10K. Figure \ref{fig:v_invertida} shows this relation between the supervised and self-supervised models ImageNet-1K classification accuracy and the Spearman correlation.

It is interesting to note that there is an inverse relation, for the supervised models, between classification accuracy and the maximum correlation they obtained. It implies that the simpler the models, the better correlation. However, for the self-supervised models, there is a direct relation between classification accuracy and correlation. 

\section{Conclusion}

In this work, we explore the perceptual properties of deep learning models. We restrict ourselves to already trained models by third-party people in order to avoid dependence on architecture design or training procedures. Following this idea, we analyze the perceptual properties using two classical image quality databases already accepted by the community \cite{PONOMARENKO_tid} and \cite{kadid10k}.  

Multiple factors are analyzed and conclusions can be raised for all of them: 

\begin{itemize}
    \item Different models: All models are better than SSIM on prediction of human correlation. Simpler models have better perceptual behavior than complex models and are better than usual image quality metrics.
    
    \item  Training objectives: While all training objectives have good perceptual properties, there are some that have better properties than others. In particular, the best results are obtained by the model trained for classification.

    \item  Amount of training data: Differences in the amount of data used for training didn't have a big effect on perceptual behavior, but training the models with over a million big natural images seems to be beneficial.

    \item  Measure concatenating outputs or using statistical descriptor: Concatenating outputs of different layers (as in LPIPS \cite{lpips} and DISTS \cite{dists}) or just taking the output of one layer does not have a big effect on the correlation. The use of statistical descriptors as the channel mean, channel mean and standard deviation or the Gram Matrix has a bad effect in the correlations.

    \item  Fine-tuning the measures: Fine-tuning the channel relevance  for a particular image quality database seems to have a negative effect when measuring correlation in a different database. A relevant output is that even without fine-tuning, the correlation of already trained models surpasses most image quality metrics. In fact, they surpass LPIPS \cite{lpips} and DISTS \cite{dists}. 

    \item   The best behavior is obtained by the fifth layer of the original AlexNet model: trained for classification, without using fine-tuning on perceptual data, without concatenating outputs, and without using statistics.
\end{itemize}

{\small
\bibliographystyle{ieee_fullname}
\bibliography{egbib}
}

\end{document}